\documentclass[letterpaper, 10 pt, conference]{ieeeconf}

\IEEEoverridecommandlockouts                              

\usepackage[utf8]{inputenc}
\usepackage{multirow}
\usepackage{graphicx}
\usepackage{subcaption}
\usepackage{amsmath}

\usepackage{xcolor}

\definecolor{mygreen}{RGB}{0, 180, 0}

\title{\LARGE \bf
	Human-Inspired Soft Anthropomorphic Hand System for Neuromorphic Object and Pose Recognition Using Multimodal Signals 
}

\author{Fengyi Wang$^{1}$, Xiangyu Fu$^{1}$, Nitish Thakor$^{2}$ and Gordon Cheng$^{1}$ 	
	\thanks{$^{1}$Fengyi Wang,  Xiangyu Fu, and Gordon Cheng are with the Institute for Cognitive Systems, Technical University of Munich, Arcisstrae 21, 80333 Munich, Germany {\tt\small \{ fengyi.wang, xiangyu.fu, gordon \} @tum.de}
	} %
	\thanks{$^{2}$Nitish Thakor is with the Department of Biomedical Engineering, Johns Hopkins University, Baltimore, MD, USA {\tt\small thakorjhu@jhu.edu}}
}
\begin{document}
	\maketitle

	\begin{abstract}
		The human somatosensory system integrates multimodal sensory feedback, including tactile, proprioceptive, and thermal signals, to enable comprehensive perception and effective interaction with the environment. Inspired by the biological mechanism, we present a sensorized soft anthropomorphic hand equipped with diverse sensors designed to emulate the sensory modalities of the human hand. This system incorporates biologically inspired encoding schemes that convert multimodal sensory data into spike trains, enabling highly-efficient processing through Spiking Neural Networks (SNNs). By utilizing these neuromorphic signals, the proposed framework achieves 97.14\% accuracy in object recognition across varying poses, significantly outperforming previous studies on soft hands. Additionally, we introduce a novel differentiator neuron model to enhance material classification by capturing dynamic thermal responses. Our results demonstrate the benefits of multimodal sensory fusion and highlight the potential of neuromorphic approaches for achieving efficient, robust, and human-like perception in robotic systems. 
	\end{abstract}

	\section{Introduction}
	
	The human somatosensory system integrates a wide range of neural inputs from different modalities to construct a comprehensive perception of the body and its interaction with the environment. Among these modalities, tactile sensation, proprioception, and thermal sensations play essential roles. Tactile sensation in the skin is mediated by various types of mechanoreceptors with distinct dynamic properties. The collective response of mechanoreceptor populations to different tactile stimuli enables the perception of fine textures, pressure, and vibration \cite{abraira2013sensory}. Beyond touch, proprioceptors located in muscle spindles provide continuous feedback on the position and movement of body parts \cite{gibsonObservationsActiveTouch1962}, \cite{ledermanExtractingObjectProperties1993}. Simultaneously, thermoreceptors detect temperature variations, which are crucial for material differentiation.

	The interplay of these multimodal sensory signals at higher levels of the perceptual pathway allows humans to perceive a wide range of object properties, including three-dimensional structures, weight distributions, and thermal properties. Furthermore, this integration facilitates the association of sensory inputs with abstract concepts such as object type and its material.
	
	Inspired by the higher-level perception and information processing mechanisms of humans, Spiking Neural Networks (SNNs), which closely mirror the communication mechanisms of biological neurons, have been proposed. Rather than relying on continuous, dense computations as in conventional artificial neural networks (ANNs), SNNs leverage sparse binary spike events to encode and process input signals. Their intrinsic advantages, such as noise robustness and power efficiency, particularly when implemented on neuromorphic chips \cite{orchardEfficientNeuromorphicSignal2021}, making them well-suited for robotic perception tasks, where the energy budget is considerably low.
	
	Anthropomorphic robotic hands serve as an ideal robotic platform for studying human-like perception in artificial systems. These devices are designed to emulate both the structural and functional aspects of the human hand. However, they usually only provide the state of the actuator or are only integrated with fingertip force sensors and lack multimodal sensory feedback, which is necessary for human-like perception.
	
	\begin{figure*}[!hbt]
		\centering
		\includegraphics[width=0.9\linewidth]{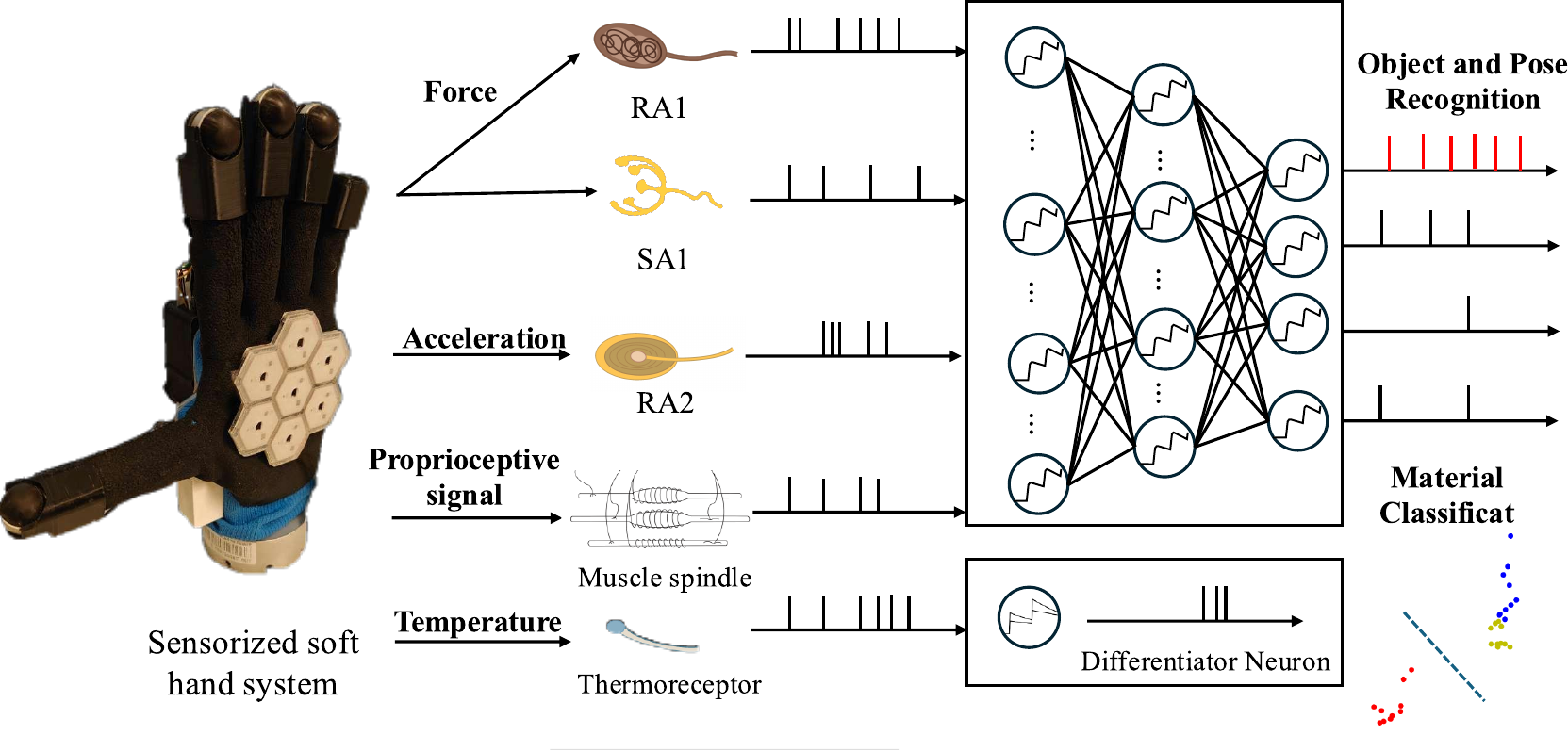}
		\caption{The schematic diagram of the human-inspired multimodal perception system. (a) The sensorized soft hand system comprises the QB SoftHand, OptoForce 3D tactile sensors, robot skin cells, and soft stretch sensors. (b) Sensory readings are encoded into spike trains using the corresponding biological receptor models. (c) The neuromorphic classifier processes the encoded signals for object and pose recognition. Thermal features generated by the thermoreceptor model and the differentiator neuron are used for material classification. 
		}
		\label{fig:system}
	\end{figure*}
	
	To address this limitation, we integrated multiple sensors with the hand to provide most of the feedback modalities that a human hand possesses. These sensors allow the system to perceive object properties such as shape, weight and thermal property during active interaction. The sensory readings are encoded into spike trains with corresponding biological receptor models to make them suitable to be processed by SNN despite the difference in modality and update frequency of the sensor. A schematic diagram of the system is shown in Fig. \ref{fig:system}. This approach not only outperforms previous studies in similar pose-invariant object classification tasks, but also achieved outstanding results in more challenging experiments involving object pose recognition. These robotic tasks facilitate the analysis of biological encoding schemes and multimodal integration in the perception pathways, while also offering practical approaches for implementing multi-sensory integration in soft robotic perception.
	
	In summary, the main contributions of this paper are:

	\begin{itemize}
		\item Integration of diverse sensors emulating the comprehensive sensory modalities of the human hand with the QB SoftHand, creating a hardware platform that advances research in haptic and thermal perception.

		\item Examine of a neural encoding scheme based on the biological model of mechanoreceptors, proprioceptors, and thermoreceptors, converting multimodal sensory signals into neuromorphic signals that can be processed by SNNs in real time.

		\item Benchmarking the performance of an SNN-based classifier that integrates neuromorphic signals by comparing its performance against state-of-the-art machine learning models in recognizing objects across multiple poses.
		
		\item Examine the use of differentiator neurons with two state variables to capture dynamic changes in temperature sensory signals. Combining the differentiator neuron signal with temperature coding enhances the system's ability to distinguish materials through their thermal properties.
		
		
		\item Demonstration of the benefits of multisensory fusion through ablation studies, highlighting the potential of neuromorphic approach in multi-modal perception for robotic applications.

	\end{itemize}

	\section{Background and Related Work}

	Robotic perception through direct interaction with objects has become a pivotal area of research, particularly in tasks such as object classification \cite{hosodaRobustHapticRecognition2010}, \cite{kerzelNeuroRoboticHapticObject2019}, \cite{khin2021hand}, manipulation \cite{jamesSlipDetectionGrasp2021}, and texture discrimination \cite{sankarTextureDiscriminationSoft2021}, \cite{taunyazovEffectiveTactileIdentification2019}.
	
	Equipping robotic devices with various sensors is a common approach to enhancing their ability to perceive objects. High accuracy in texture and object discrimination can be achieved using a robust tactile descriptor combined with intricate manipulation performed by the Shadow Dexterous Hand equipped with fingertip tactile sensors \cite{kaboliTactilebasedActiveObject2019}. This anthropomorphic hand, driven by up to twenty actuators within its joints, can perform nearly all human gestures and even some beyond human capability. Furthermore, it provides precise joint position data as proprioceptive feedback, enabling easy determination of the hand's configuration using forward kinematics. However, its operation needs prior knowledge of the object and intricate planning to conform to it.
		
	In contrast, soft robotic hands typically incorporate synergy mechanisms similar to human hands \cite{catalano2014adaptive}, and can almost automatically conform to the grasped object.  However, these systems generally lack proprioceptive feedback and usually rely solely on tactile sensing for perception tasks \cite{kerzelNeuroRoboticHapticObject2019, rostamianTextureRecognitionBased2022}. Consequently, their compliance is not fully utilized for perceiving object shapes, and they frequently struggle to achieve high accuracy in pose-invariant object recognition tasks. In classification tasks involving 10–20 objects, their accuracy typically remains below 90\% \cite{khin2021hand}, and under specific setups, it can drop as low as 50\% \cite{kerzelNeuroRoboticHapticObject2019}.
	
	In human somatosensory systems, tactile sensation and proprioception play essential roles in perceiving and interacting with objects. Tactile perception arises from various mechanoreceptors in the skin \cite{abraira2013sensory}, while proprioceptive feedback, mediated by muscle spindles, encodes the position and movement of body parts through spiking neural activity \cite{hulligerMammalianMuscleSpindle1984, matthewsEvolvingViewsInternal1981}. These two sensory modalities are seamlessly integrated within the somatosensory system to derive various properties of objects during active manipulation, such as weight distribution, texture, and shape.
	
	Furthermore, humans utilize the ability to perceive thermal properties for object recognition, as it offers sensory cues that complement tactile features. For instance, humans can differentiate materials, such as metal and wood, based on how quickly they conduct heat to the skin. The thermal feedback, mediated by thermoreceptors in the skin, is particularly valuable when other sensory inputs, such as texture or weight, are ambiguous. Integrating this capability into soft robotic systems could significantly enhance their ability to classify and interact with objects, mimicking the multimodal sensory integration observed in humans. However, although thermal properties have been successfully utilized to differentiate various objects and are considered effective \cite{kaboliTactilebasedActiveObject2019}, few soft anthropomorphic hands incorporate this information into such tasks.
	
	Sensorized soft anthropomorphic hand systems are ideal for investigating multisensory feedback in artificial systems inspired by human perception. Most previous research primarily focused on pose-invariant object classification tasks using tactile information. Our prior work attempted to address similar tasks using human-like proprioceptive signals \cite{fengyiObject}, although achieved accuracy was suboptimal. In this work, we aim to develop methods that go beyond object classification and also determine the poses of objects.
	
	To enable the system to have a more comprehensive understanding of the object, we need to integrate multiple sensors with different modalities into a single system. However, these sensors typically operate at different update rates asynchronously. They often fail to produce synchronized, uniform-length data matrices that popular machine learning algorithms require. Interpolation is a possible workaround to generate such data, but it introduces additional computational overhead, which increases both system response time and hardware performance demands. Designing a synchronized multimodal sensor system requires considerable effort. Moreover, continuously updating large amounts of sensor data can exhaust transmission bandwidth in large-scale systems. This limitation presents significant challenges for large-scale applications, particularly when compared to event-driven systems like robotic skin \cite{cheng2019comprehensive}.
	
	Neuromorphic encoding methods and spiking neural networks (SNNs), inspired by the human neural system, present a promising solution to these challenges. Converting sensory readings into spike trains through biologically inspired receptor models provides an efficient representation of sensory information in the time domain. This encoding strategy is well-aligned with the sparse and asynchronous processing characteristics of spiking neural networks, which closely emulate the information processing mechanisms of biological neurons. Unlike traditional clock-driven systems, event-driven approaches transmit and process data only when specific events occur, minimizing redundant data transmission and reducing computational overhead. Moreover, this approach is highly compatible with neuromorphic hardware, leveraging their specialized architectures to enhance power efficiency, scalability, and performance in large-scale robotic and sensory systems \cite{orchardEfficientNeuromorphicSignal2021}.

	\begin{figure}
		\centering
		\includegraphics[width=1\linewidth]{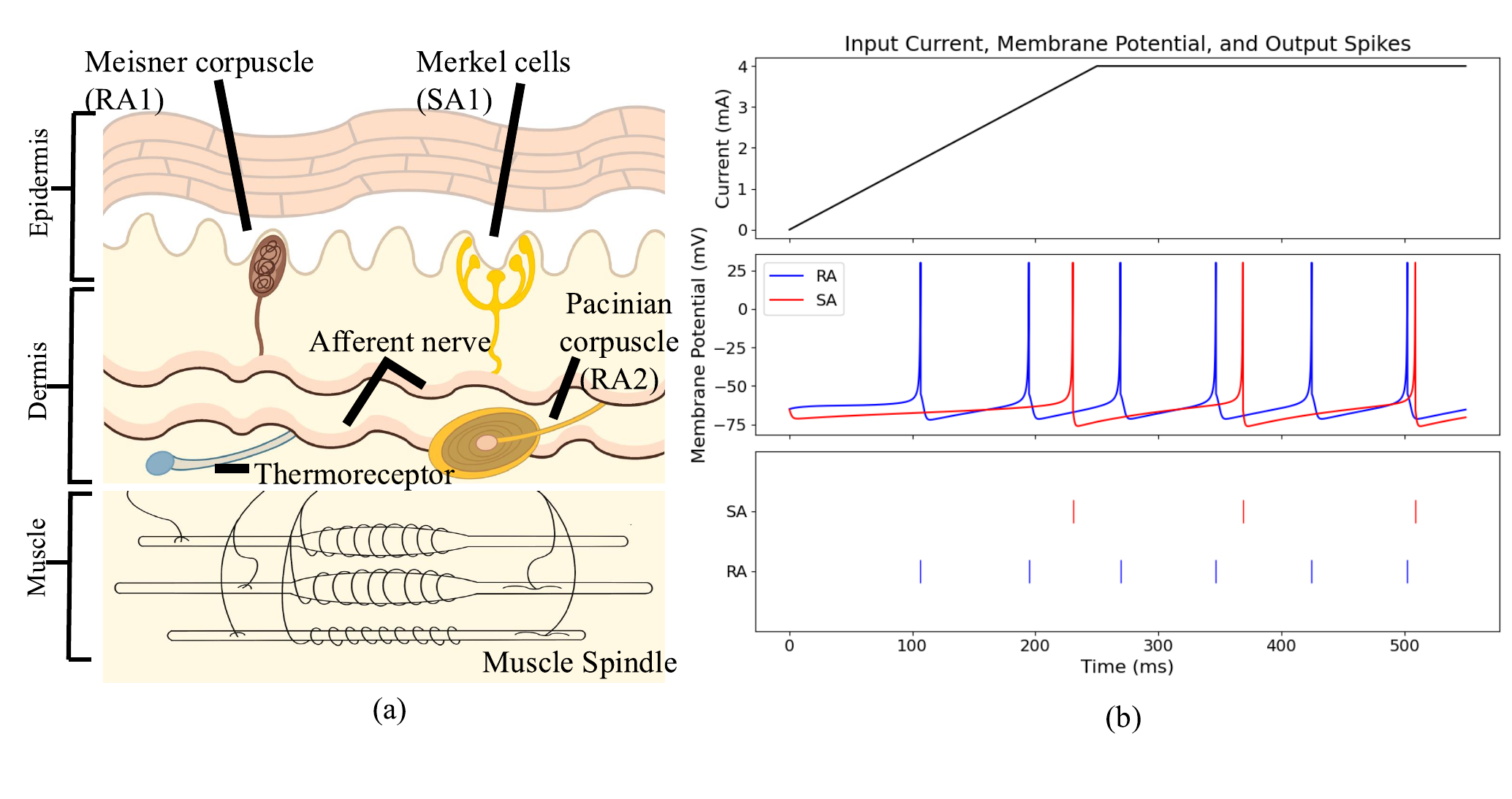}
		\caption{(a) illustrates the mechanoreceptors, thermoreceptors, and muscle spindles embedded in human skin and muscles, responsible for encoding tactile, thermal, and proprioceptive sensory feedback. (b) depicts the spiking responses of SA and RA mechanoreceptors. SA mechanoreceptors respond to static forces, continuing to fire during constant stimulation, whereas RA mechanoreceptors are more sensitive to dynamic changes, firing also during the onset of the applied force.}
		\label{fig:humanskin}
	\end{figure}
	
	\section{Methods}

	\subsection{The Sensorized Soft Anthropomorphic Hand}
	 \label{mtd:sensors}
	In this work, we employed the QB SoftHand, a soft robotic hand inspired by the synergy mechanism in grasping design \cite{catalano2014adaptive}. Like humans, who can effortlessly conform their grasp to objects with minimal cognitive effort, the QB SoftHand achieves various grasping actions through its soft and adaptive synergy design without the need for explicit joint-level control. However, the QB SoftHand lacks sensory feedback beyond motor position and current, lacking multimodal feedback in human skin and muscle, such as force, proprioception, and temperature. The absence of necessary sensory feedback makes it infeasible to infer precise finger configurations, not to mention extracting other object properties.
	
	To address these limits, we installed various sensors of different modalities on different parts of the robotic hand to provide the essential sensory feedback necessary for human-like perception.
	
	\subsubsection{3D Force Sensors} Inspired by the sensitivity of human fingertips, we installed the Optoforce sensor set, which includes four OMD-20-SE-40N 3D tactile sensors with 3D-printed finger cots at the distal ends of four fingers of the QB SoftHand, excluding the little finger, to enhance its sensory capabilities. These tactile sensors are capable of measuring forces along three axes by utilizing infrared light to detect small deformations in their outer surface, with a sampling rate of 350 Hz.
	
	\subsubsection{Proprioceptive Sensors}  Human proprioception primarily relies on muscle spindle signals that encode the length of the muscle and its rate of change. Therefore, we installed five stretch sensors on the dorsal side of the QB SoftHand's fingers. An enclosure fixes one end of a sensor on the back of the hand and the other by a finger cot on the fingertip. similar to the approach used in our prior work \cite{fengyiObject}.
	
	An ESP32 microcontroller, placed in the enclosure on the dorsal side of the hand, reads the stretch sensor measurements through a TCA9548 multiplexer at 150 Hz and encodes them into neuromorphic signals.
	
	\subsubsection{Robot Skin} Although less sensitive than the fingertips, the skin on the human palm perceives various types of information, including force, acceleration, and temperature. To mimic this functionality, the palm of the QB SoftHand was covered with seven robot skin cells. Each robot skin cell is a sensing unit comprises three normal force sensors, two temperature sensors, and one three-axis accelerometer.
	
	The event-driven communication architecture allows the robotic skin cells to transmit multimodal sensing information at 250 Hz on a large scale \cite{cheng2019comprehensive}. However, this architecture introduces challenges for online processing in conventional artificial neural networks (ANNs) due to asynchronous updates and irregular intervals between skin cell signals. 
		
	The sensorized soft hand system is illustrated in Fig. \ref{fig:system}(a), and the sensory feedbacks provided by the hand are summarized in Table \ref{tab:sensors}.
	

	\begin{table}[b]
		\centering
		\caption{Sensing Capability of the sensorized hand}
		\label{tab:sensors}
		\resizebox{\columnwidth}{!}{%
			\begin{tabular}{cccccc}
				Sensor & Position & Modality & \begin{tabular}[c]{@{}c@{}}Channels per \\Sensor\end{tabular} & \begin{tabular}[c]{@{}c@{}}Number of \\ Sensors\end{tabular} & \begin{tabular}[c]{@{}c@{}}Total \\Channels\end{tabular}  \\ \hline
				Optoforce Sensor & Fingertip & Force & 3 & 4 & 12 \\ \hline
				&  & Force & 3 & 7 & 21 \\ \cline{3-6} 
				Robot skin cell & Palm & Acceleration & 3 & 7 & 21 \\ \cline{3-6} 
				&  & Temperature & 2 & 7 & 14 \\ \hline
				Strech sensor & \begin{tabular}[c]{@{}c@{}}Dorsal side \\ of finger\end{tabular} & Stretch & 2 & 5 & 10 \\ \hline
				Total &  &  &  &  & 78
			\end{tabular}%
		}
	\end{table}
	
	\subsection{Neuromorphic Encoding Scheme}
	\label{subsec:encoding}

	\subsubsection{Force}
	Tactile sensation in glabrous skin is primarily mediated by mechanoreceptors in the epidermis: Meissner corpuscles, classified as Rapidly Adapting (RA) receptors, and Merkel cells, classified as Slowly Adapting (SA) receptors, as illustrated in Fig. \ref{fig:humanskin}.
	
	Inspired by human mechanoreceptors, sensory readings from the Optoforce 3D tactile sensors and the robotic skin cells were encoded into spiking activities by mimicking the SA and RA mechanoreceptors. This encoding scheme was implemented using the Izhikevich model \cite{izhikevichSimpleModelSpiking2003} with corresponding parameters. The dynamics of the neuron model are governed by Equations (\ref{Eq:Iz1}), (\ref{Eq:Iz2}), and (\ref{Eq:Iz3}).
	
	\begin{equation}\label{Eq:Iz1}
		\frac{d v}{d t}=0.04 v^2+5 v+140-u+I
	\end{equation}
	
	\begin{equation}\label{Eq:Iz2}
		\frac{d u}{d t}=a(b v-u)
	\end{equation}
	
	\begin{equation}\label{Eq:Iz3}
		\text { if } v \geq \vartheta, \text { then }\left\{\begin{array}{l}
			v \leftarrow c \\
			u \leftarrow u+d
		\end{array}\right.
	\end{equation}
	where $v$ is the membrane potential, $u$ is the recovery variable, $I$ is the input current, The parameters $a$, $b$, $c$, $d$ are Izhikevich parameters that control neuron dynamics. The values of these parameters  are shown in Table \ref{tab:force-params}.

	\begin{table}[ht]
		\centering
		\caption{Encoding Neuron Parameters}
		\label{tab:force-params}
		\resizebox{0.7\columnwidth}{!}{%
			\begin{tabular}{cccccc}
				& \textbf{a} & \textbf{b} & \textbf{c} & \textbf{d} & $\vartheta$ \\ \hline
				\textbf{Izhikevich SA} & 0.02       & 0.2        & -65        & 8  & 30               \\
				\textbf{Izhikevich RA} & 0.02              & 0.25       & -55         & 4     & 30                      
			\end{tabular}%
		}
	\end{table}

	\subsubsection{Acceleration}
	
	In the human sensory system, high-frequency vibrations that often indicate the onset or offset of touch are encoded by Pacinian corpuscles. They are RA mechanoreceptor lying in the deeper layer of human skin as shown in Fig. \ref{fig:humanskin}. Inspired by this mechanism, we utilize three Izhikevich RA neurons to encode each three-axis acceleration reading provided by the robotic skin.
	
	\subsubsection{Proprioceptive Signals}
	
	In this work, we use an ESP32 to read the soft stretch sensor measurements and encode them in real time into spike trains. The encoding mechanism for proprioceptive signals is adapted from our prior work \cite{fengyiObject} based on Vannuci's neuromorphic muscle spindle model \cite{vannucciProprioceptiveFeedbackNeuromorphic2017}. 
	This mechanism encodes the relative length of the soft stretch sensor,
	$L$, and its rate of change, $\dot{L}$, into the spiking activities of primary (Ia) and secondary (II) afferent neurons using a homogeneous Poisson process \cite{heeger2000poisson}. It has demonstrated efficacy in object classification tasks based on proprioceptive signals.

	\subsubsection{Temperature}
	To mimic the firing pattern of the biological heat-sensitive neurons \cite{wang2018sensory}, we adopted the parameters of the Izhikevich RA model with half Gaussian-like tuning curve \cite{wang2023bio}. The input current of the thermoreceptor model follows equations (\ref{eq:tuning1}) and (\ref{eq:tuning2}):
	
	\begin{equation}
		\label{eq:tuning1}
		I=\exp \left(-\frac{1}{2} \frac{(T_{max} - T)^{2}}{\sigma^{2}}\right) \cdot I_{max} 
	\end{equation} 
	\begin{equation}
		\label{eq:tuning2}
		\textrm{if\ } T \geq T_{max},\  \textrm{then}\  I = I_{max}
	\end{equation} 
	where $T$ represents the temperature reading of the robot skin sensor, and $T_{max}$ is set to $52\ ^{\circ}$C at which the heat-sensitive neuron is fully activated according to the activation threshold of the TRPV2 ion channel. We set $I_{max}$ to 20 mV/ms, and $\sigma = T_{max} - T_a$ for heat-sensitive neuron with activation temperature $T_a$, so that the step input with magnitude of $T_a$ results in burst activity. The neuron spikes regularly as the ongoing activity in the biological heat-sensitive neurons at lower temperature \cite{darian1979warm}. 
	
	In this work, we only implemented one population of neurons with $T_a = 38\ ^{\circ}$C, since the other populations with higher $T_a$ are highly inactive under the setup of this work.

	\begin{figure}
		\centering
		\includegraphics[width=\linewidth]{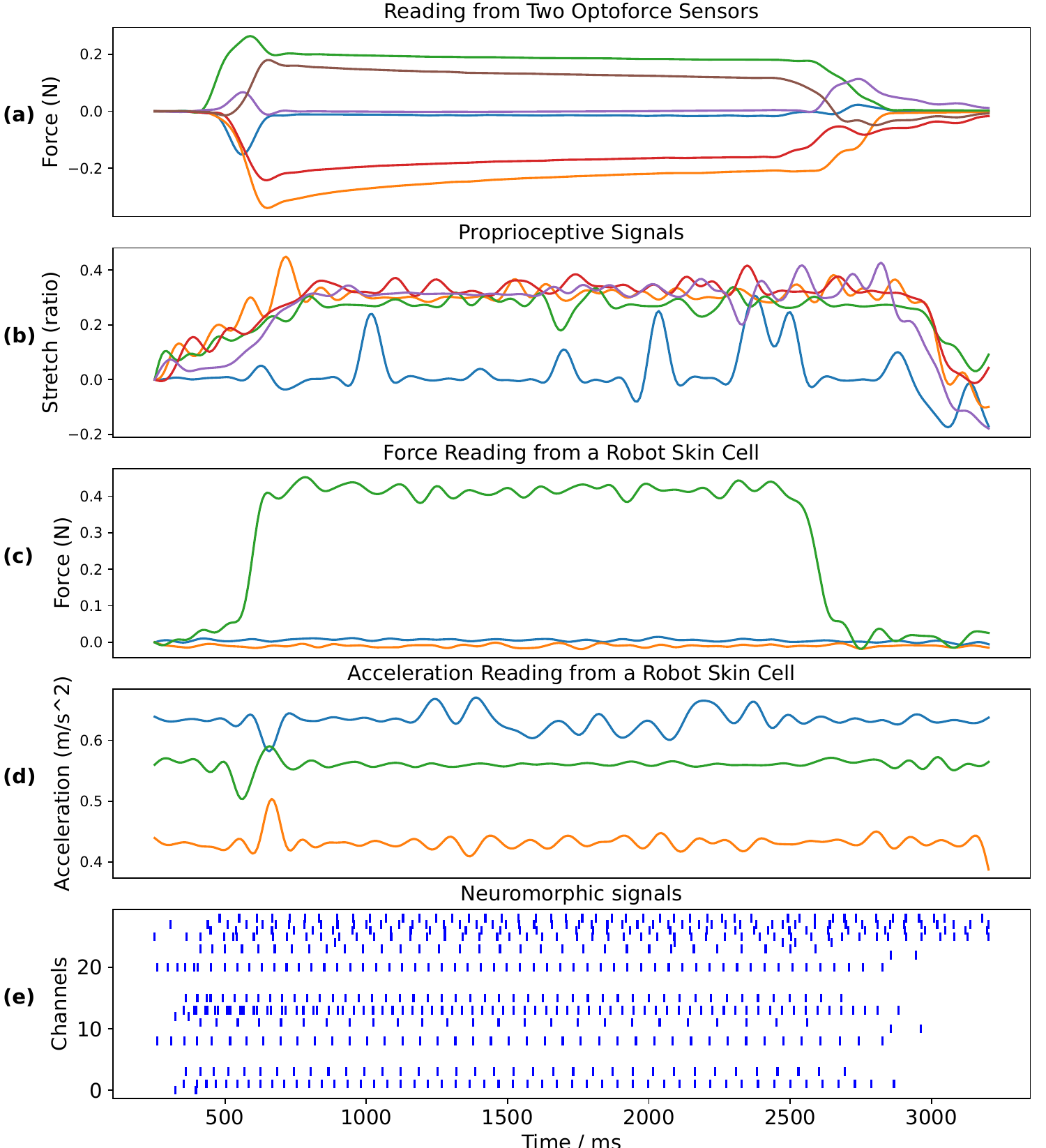}
		\caption{(a) 3D fingertip force readings from two Optoforce sensors, (b) proprioceptive readings from the stretch sensors on the dorsal side of the fingers, (c) normal force readings from a robot skin cell, (d) acceleration readings from a robotic skin cell, (e) encoded neuromorphic signals.}
		\label{fig:code}
	\end{figure}

	\subsection{Neuromorphic Classifier for Object and Pose Recognition}

	Although Izhikevich neurons can better emulate the spiking patterns of biological neurons, it is more challenging to deploy them on neuromorphic hardware like Loihi 2 due to the the amount of variables they need. In contrast, CUrrent-BAsed Leaky Integrate and Fire (CUBA-LIF) neurons are easier to implement and offer a higher degree of nonlinearity than simple LIF neurons, enabling improved performance in tasks such as object classification. The dynamic of a CUBA-LIF neuron follows equations (\ref{Eq:cuba1}), (\ref{Eq:cuba2}), and (\ref{Eq:cuba3}).
	
	\begin{equation}
		\label{Eq:cuba1}
		\frac{d u}{d t}=-\alpha_u u +  x
	\end{equation}
	\begin{equation}
		\label{Eq:cuba2}
		\frac{d v}{d t}=-\alpha_v v+u
	\end{equation}
	\begin{equation}
		\label{Eq:cuba3}
		\text { if } v \geq \vartheta, \text { then }\left\{\begin{array}{l}
			v \leftarrow 0 \\
			s \leftarrow 1
		\end{array}\right. \text{, otherwise } s \leftarrow 0
	\end{equation}
	where $u$ is the synaptic current, $\alpha_u$ is the current decay factor, $x$ is the input spike signal, weighted by synaptic strength, $v$ is the membrane potential, $\alpha_v$ is the decay factor for the membrane potential, and $\vartheta $ is the firing threshold.
	
	The classifier used in this work is a five-layer SNN. The input to the SNN is encoded spike trains as described in Section \ref{subsec:encoding}. The four hidden layers contain 128, 256, 512, and 64 CUBA-LIF neurons, respectively. The output layer consists of CUBA-LIF neurons corresponding to the number of classes. The classification is determined based on the index of the output neuron that emits the highest number of spikes.
	
	The conventional backpropagation algorithm is not applicable to training SNNs due to the non-differentiable nature of spike functions. Recent advances, such as the Spike Layer Error Reassignment in Time (SLAYER) algorithm \cite{shresthaSLAYERSpikeLayer2018}, adopt a temporal credit assignment approach that enables error backpropagation to preceding layers for updates. The error is calculated as the difference between the output spike counts of neurons representing the correct class (positive spike counts, PSC) and other classes (negative spike counts, NSC), relative to the desired values \cite{LavaSoftwareFramework}. $\tau_{grad}$ and $\kappa_{grad}$ are parameters that control time constant and scale of spike function derivative in the algorithm. 
	
	The parameters we used for the CUBA-LIF neurons and the SLAYER training algorithm are summarized in Table. \ref{tab:snn}.

	\begin{table}[htb]
		\centering
		\caption{SNN Parameters}
		\label{tab:snn}
		\resizebox{\columnwidth}{!}{%
			\begin{tabular}{ccccccc}
				\multicolumn{3}{c}{Neuron parameters} & \multicolumn{4}{c}{Training parameters} \\ \hline
				$\vartheta$ & $\alpha_u$ & $\alpha_v$ & PSC ratio & NSC ratio & $\tau_{grad}$ & $\kappa_{grad}$ \\
				1.25 & 0.25 & 0.03 & 0.5 & 0.1 & 0.03 & 3
			\end{tabular}%
		}
	\end{table}

	In this work, the SNN is trained and tested on a conventional GPU. Alternatively, the model can be deployed on neuromorphic hardware for execution in a fast and power-efficient manner.
	
	
	\subsection{Material Classification with Differentiator Neuron}
	\label{sec:diff_neuron}
	The mean inter-spike interval (MISI) is a widely used metric in neuromorphic classification tasks \cite{iskarousScalableAlgorithmBased2021a}. However, it does not always capture the difference in spike dynamics, as the MISI can remain the same even when the the temperature curves  dynamics differ. To address this problem, we propose a differentiator neuron characterized by two variables, $v_{fast}$ and $v_{slow}$. The dynamics of both variables can be described by equations (\ref{Eq:diff}) to (\ref{Eq:diff3}).
	
	
		\begin{equation}\label{Eq:diff}
			\frac{d v}{d t}=-\frac{v}{\tau}
		\end{equation}
	

	\begin{equation}
		 v\longleftarrow v(t)+w \cdot s_i.
	\end{equation}	
	\begin{equation}\label{Eq:diff3}
		\text { if } v_{fast} - v_{slow} \geq \vartheta_{diff}, \text { then }\left\{\begin{array}{l}
			v_{fast} \leftarrow 0 \\
			v_{slow} \leftarrow 0 \\
			s_o \leftarrow 1
		\end{array}\right.
	\end{equation}
	where $\tau$ is the decay factor, $s_i$ is the input spike, and $s_o$ is the output spike. The differentiator neuron is implemented with the Brian2 framework \cite{brian2}, and the parameters used in this work are listed in Table \ref{tab:Differentiator Neuron}.
	
\begin{figure}[htb]
		\centering
		\includegraphics[width=1\linewidth]{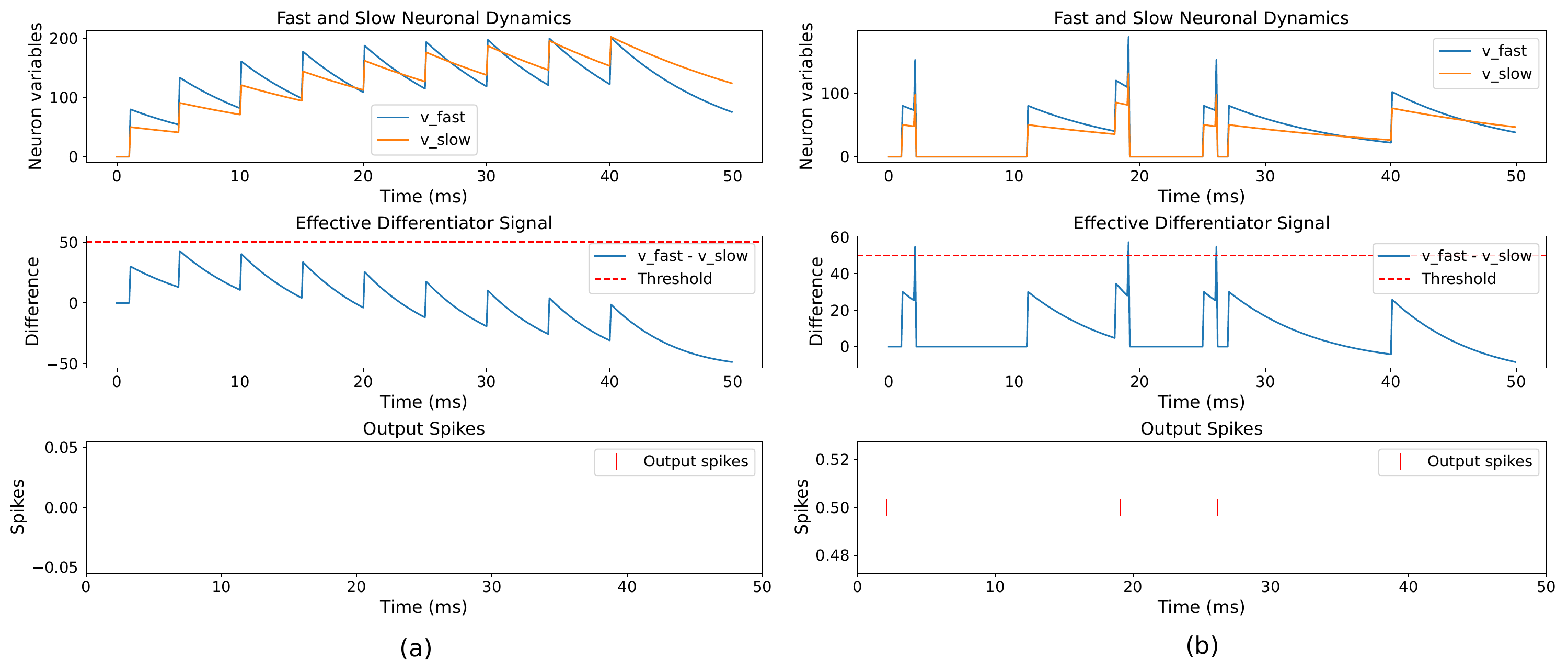}
		\caption{The response of a differentiator neuron to distinct input spike trains, each containing the same number of input spikes. In (a), the neuron, when stimulated by a consistent 200 Hz spike train, does not generate any output spikes. Conversely, in (b), the neuron produces output spikes in response to the irregular spike train.}
		\label{fig:neurondynamic}
	\end{figure}

	Fig. \ref{fig:neurondynamic} illustrates how the differentiator neuron responds to varying spike rates. The MISI of the differentiator neuron serves as a measure of material-specific spike responses, highlighting distinct features for different materials.

	\begin{table}[hb]
		\centering
		\caption{Differentiator Neuron Parameters}
		\label{tab:Differentiator Neuron}
		\resizebox{0.8\columnwidth}{!}{%
			\begin{tabular}{ccccc}
				$w_{slow}$ & $w_{fast}$ & $\tau_{slow}$ & $\tau_{fast}$ & $\theta_{diff}$ \\ \hline
				50 mV & 80 mV & 50ms & 30 ms & 50 mV
			\end{tabular}%
		}
	\end{table}

	\section{Experiment}

\begin{figure}[ht]
		\centering
		\includegraphics[width=0.9\linewidth]{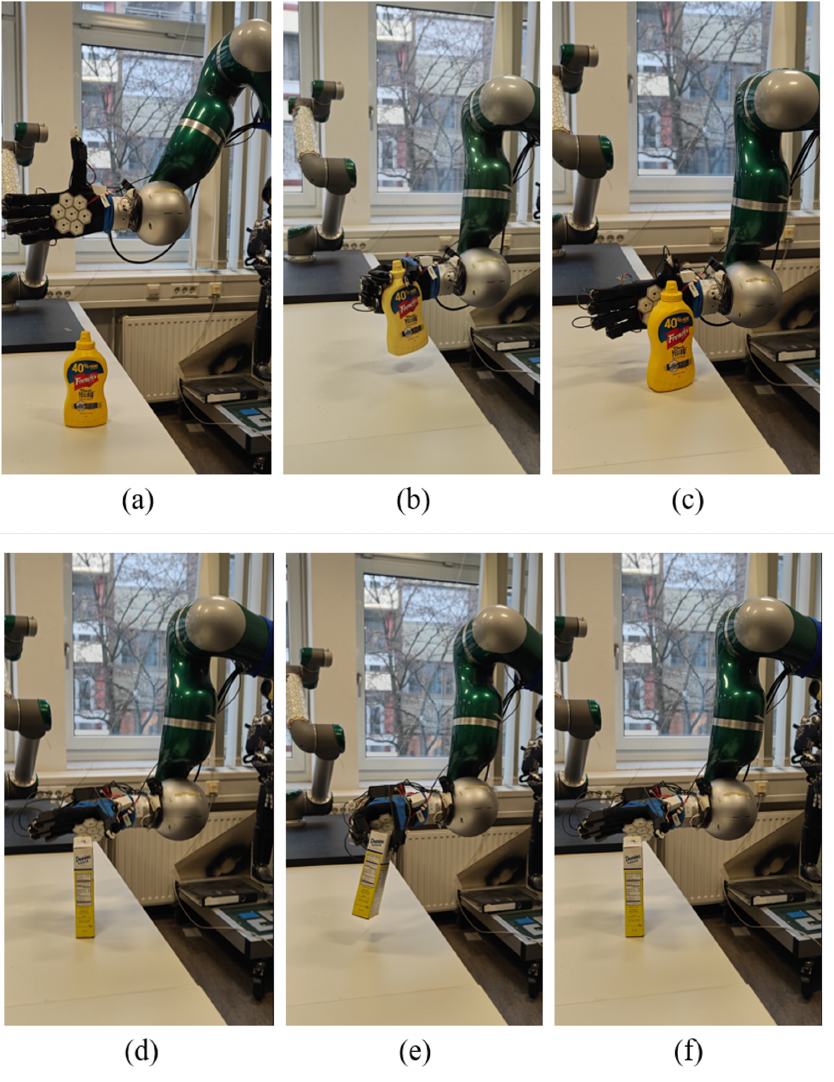}
		\caption{Illustration of data collection process for the object and pose recognition task. (a) depicts the initial setup. (b) represents the grasp-lift stage of the exploration. In (c), the object is released in preparation for the next trial. (d), (e), and (f) illustrate the corresponding actions during a top grasp.}
		\label{fig:grasp_process}
	\end{figure}

	The sensorized QB SoftHand is mounted on the 7-DOF KUKA LBR iiwa robotic arm, as illustrated in Fig. \ref{fig:grasp_process}. The entire system is operated via the Robot Operating System (ROS), which facilitates precise control, autonomous movement, and data processing.
	
	We first evaluated the performance of the SNN-based classifier in an object and pose recognition task. The classifier utilizes neuromorphic signals provided by various human receptor-inspired models. Additionally, we assessed the efficacy of multisensory fusion through ablation tests. To eliminate the influence of varying ambient temperature over time in the data collection process, temperature reading was discarded from the object and pose recognition task. We conducted a separate material classification task to demonstrate the enhanced perception capability of the sensorized soft hand system.

	\subsection{Object and Pose Recognition Tasks}
	
	In the data collection process, the robotic hand approached objects  laterally and from above. For lateral grasping, the objects were placed in two distinct poses: one with the object facing the robotic hand and the other after rotating the object by 90 degrees. Across all trials, the sequence of actions: approaching, grasping, lifting, and releasing remained consistent. The objects were not restored after release and was not necessarily identical to its initial pose before being grasped.
	
\begin{figure}
		\centering 
		\includegraphics[width=1\linewidth]{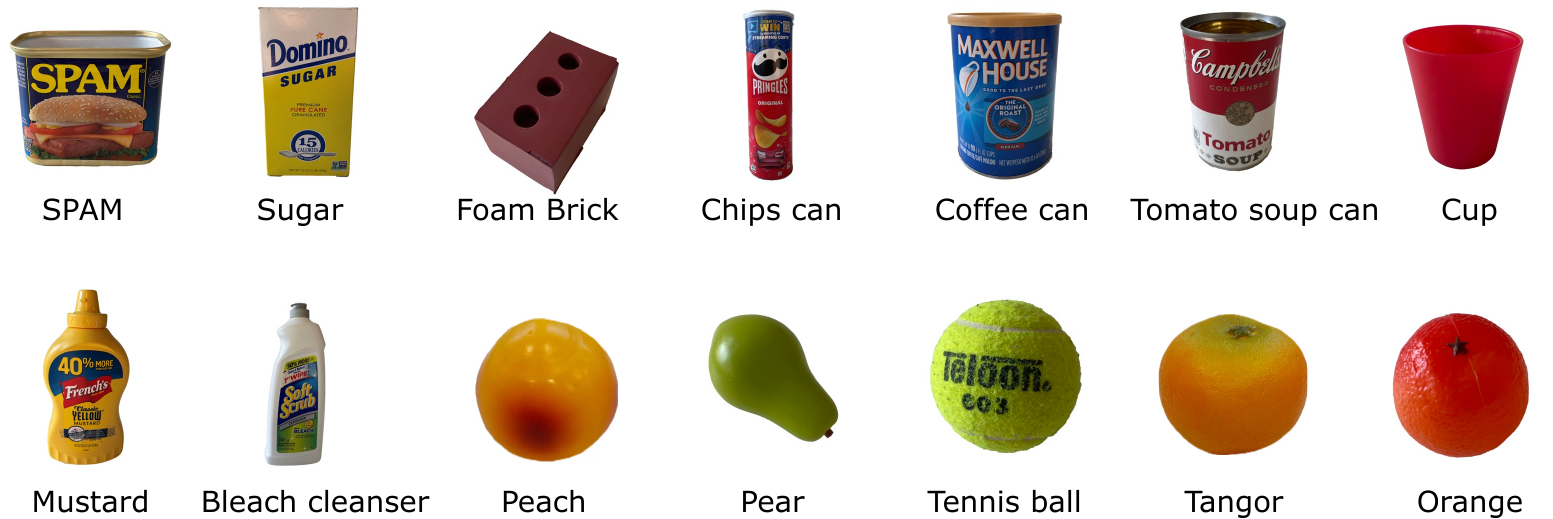}
		\caption{The objects from the YCB benchmark were used for the object and pose recognition dataset collection. This dataset includes objects with similar geometry, such as oranges and tangors, as well as soft objects like the foam brick and tennis ball.}
		\label{fig:datasetwithnames}
	\end{figure}

	Fourteen objects from the YCB benchmark dataset were used, with 30 trials performed consecutively for each pose. During each trial, data was recorded from the initiation of the approaching movement until the completion of the release action. Fig. \ref{fig:datasetwithnames} shows the objects used in the experiments.
	
	The dataset was split into training and test sets using a 70/30 ratio. Each model was trained for a maximum of 1000 epochs and tested on the test set. This process was repeated five times, and the mean accuracy and standard deviations of different classifiers are summarized in Table \ref{tab:acc}.
	
	We first conducted an 14-class object recognition task focusing solely on distinguishing object categories, followed by an object and pose recognition task aimed at differentiating both object categories and the poses of the objects.
	
	In the object and pose recognition task, we do not distinguish between the front side and left side for strictly rotationally symmetric objects, including the chip can, coffee can, tomato soup can, cup, and tennis ball, resulting in a 37-class recognition task.
	
	To assess the contribution of different modalities to the performance and to validate the assumption that integrating multiple sensor modalities improves classification accuracy, we conducted an additional ablation tests. These tests evaluated the model's performance with various sensor combinations in the object and pose recognition task.

	\subsection{Material Classification Task}
\begin{figure}[tb]
		\centering 
		\includegraphics[width=1\linewidth]{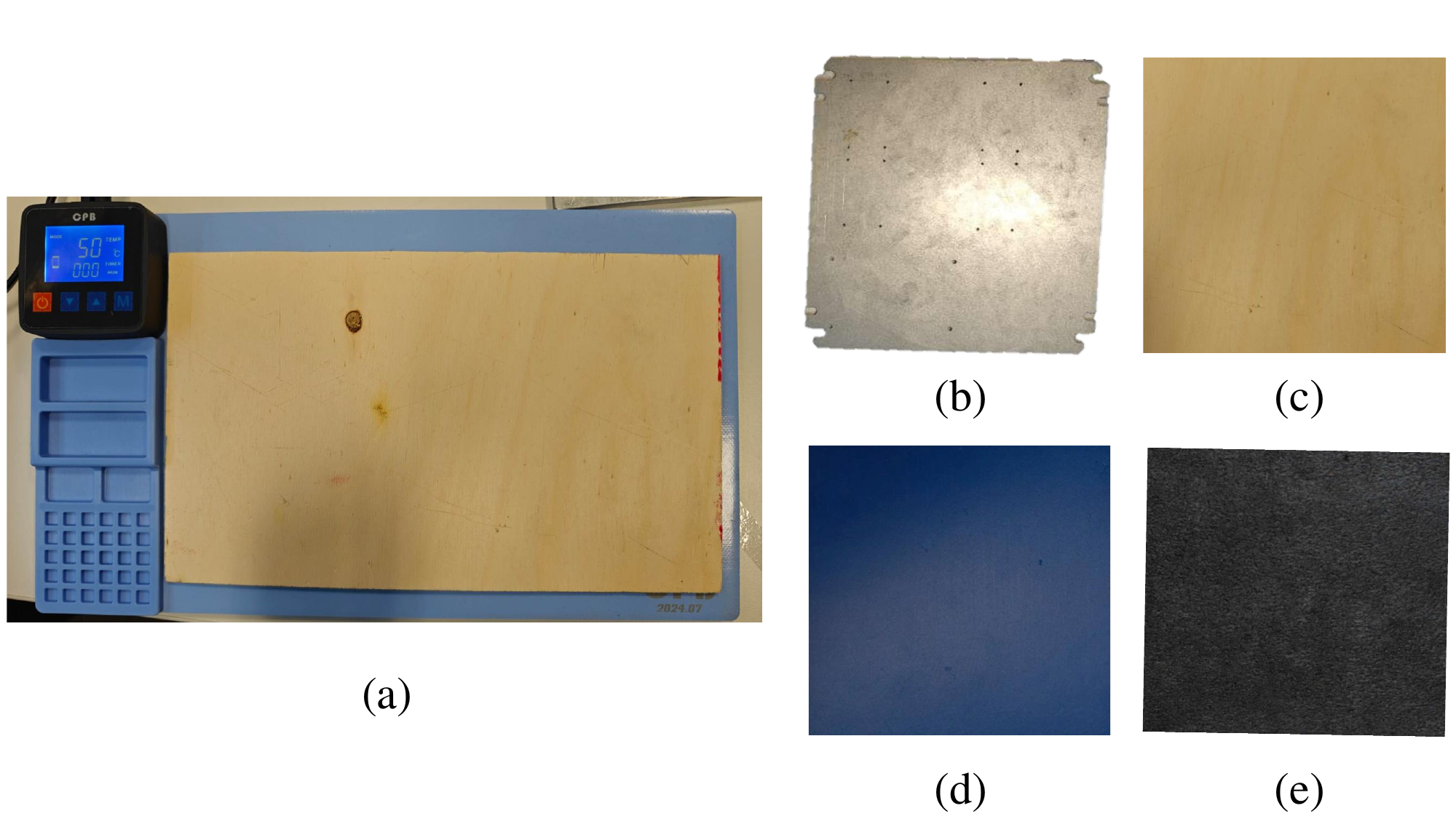}
		\caption{(a) illustrates the wooden plate sample under experimental setup. (b), (c), (d), (e) show iron plate, PVC plate, and low density polyethylene sample used in material classification task, respectively.}
		\label{fig:temp_mat}
	\end{figure}

	To validate the capability of the sensorized robotic hand in distinguishing object materials based on thermal conductivity, we conducted a static pressing experiment in which the robotic palm was pressed against the material samples shown in the Fig. \ref{fig:temp_mat}. The samples are heated evenly by a hot plate at 50 $^\circ$C. In each trail, the data collection began when the normal force detected by the robotic skin exceeded a small threshold and lasted for 30 seconds. Data collection for each material was repeated 10 times.

	\subsection{Baseline Models}
	To evaluate the performance of the proposed methods, we implemented two state-of-the-art machine learning models as benchmarks. For both models, all sensory data were interpolated to match the update frequency of the fastest sensor in the system (350 Hz). This ensured that the data input to the machine learning models was aligned in both temporal resolution and data length across all sensory modalities.
	
	\textbf{Long-Short Term Memory (LSTM):}
	We first implemented an ANN includes Long-Short Term Memory (LSTM) units \cite{hochreiterLongShortTermMemory1997}, known for their proficiency in learning and continuously predicting temporal sequences. The architecture implemented recurrent layers to process sequential data effectively. The LSTM model consists of two layers with 128 hidden units each, designed to process time-series data with 100 features per time step.
	
	\textbf{Transformer}: The Transformer model is widely known for its self-attention mechanism \cite{vaswani2017attention}, and has gained prominence for sequence processing tasks, such as natural language processing task. Its ability to handle temporal relationships via positional encoding makes it particularly suitable for the classification of object interactions based on spatial-temporal data. 
	The network employs 4 encoder layers, utilizing 4 attention heads per layer. Each encoder layer includes a feedforward network with a hidden dimension of 256.
	
	\section{Results}
	\subsection{Object and Pose Recognition Tasks}
	In both the 14-class object recognition task and the 37-class object and pose recognition task, the model utilizes neuromorphic signals converted from the tactile and proprioceptive readings from the fingertip 3D force sensors, stretch sensors, and the robotic skin. To evaluate the robustness of the models, we conducted five-fold cross-validation, and the results are summarized in Table \ref{tab:acc}. Our proposed multisensory method achieved 99.5\% accuracy in the object recognition task, significantly outperforming the results reported in previous works \cite{kerzelNeuroRoboticHapticObject2019}, \cite{khin2021hand}. Furthermore, the neuromorphic model demonstrated superior accuracy and robustness compared to baseline models in both recognition tasks despite utilizing a simple neuron model and a feed-forward structure.
		
	\begin{table}[!h]
		\centering
		\caption{Performance in Recognition Tasks}
		\label{tab:acc}
		\resizebox{\columnwidth}{!}{%
			\begin{tabular}{ccc}
				Models & \begin{tabular}[c]{@{}c@{}}Object recognition\\ (14 classes)\end{tabular} & \begin{tabular}[c]{@{}c@{}}Object and \\pose recognition\\ (37 classes)\end{tabular} \\ \hline
				\textbf{Neuromorphic classifier} & \textbf{99.50\% (0.002)} & \textbf{97.14\% (0.009)} \\
				Transformer & 94.79\% (0.012) & 92.18\% (0.019) \\
				LSTM & 92.78\% (0.015) & 86.72\% (0.090)
			\end{tabular}%
		}
	\end{table}
	\begin{table}[h]
		\centering
		\caption{Ablation Test Results}
		\label{tab:ablation}
		\resizebox{0.8\columnwidth}{!}{%
			\begin{tabular}{cc}
				Sensory Signals & Accuracy \\ \hline
				\textbf{All sensors combined }&\textbf{ 97.14\% (0.009)} \\
				Fingertip force sensors only & 89.47\% (0.008) \\
				Robot skin force only & 82.21\% (0.008) \\
				Proprioception only & 58.30\% (0.019) \\
				Robot skin acceleration only & 46.42\% (0.019) \\
				& 
			\end{tabular}%
		}
	\end{table}
	The results of ablation test on object and pose recognition task are summarized in Table \ref{tab:ablation}. When all sensory modalities were combined, the system achieved an impressive accuracy of 97.14\%. In contrast, even the best-performing single modality, the fingertip force sensors, only reached an accuracy of 89.47\%. The result clearly shows that single sensory modality is insufficient to provide the comprehensive information necessary for accurately recognizing objects and their poses.

	\subsection{Material Classification Task}
\begin{figure}[t]
		\centering
		\includegraphics[width=0.9\linewidth]{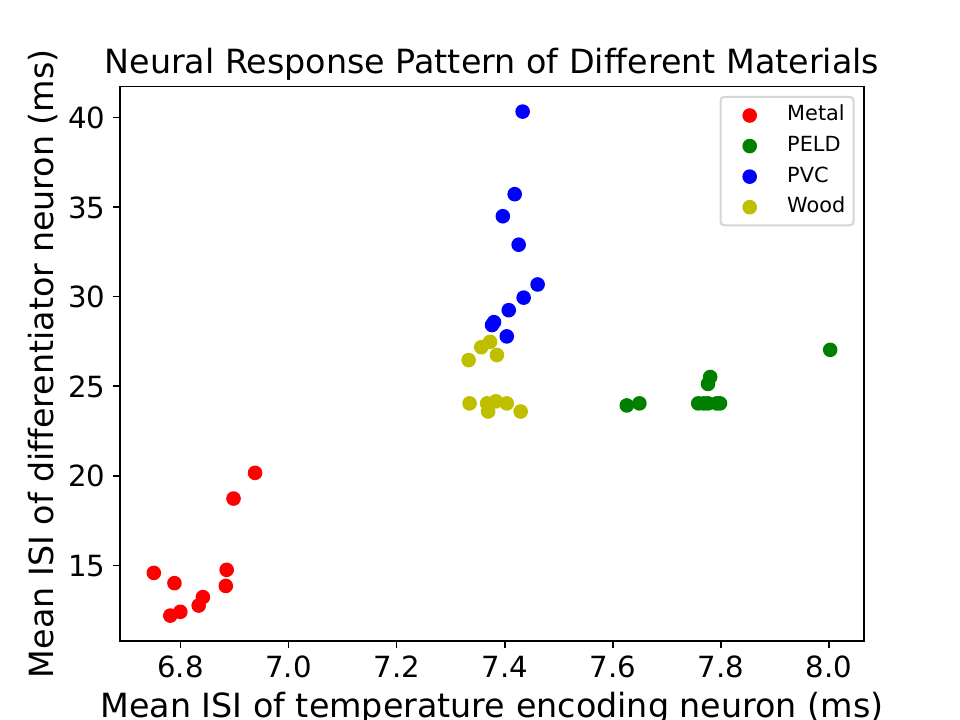}
		\caption{The thermal feature of four material samples based on the MISI of the output of thermoreceptor model and the differentiator neuron. Each trail is represented by a point on the plot, highlighting the linear separability between them.}
		\label{fig:material}
	\end{figure}

	As illustrated in Fig. \ref{fig:material}, the thermal feature combines the MISI of the thermoreceptor and differentiator output and exhibits linear separability across the tested materials. With this linear separability of the features, a Support Vector Machine (SVM) classifier trained on 70\% of the collected data can reach 100\% accuracy in five-fold cross-validation. In contrast, wood and PVC are difficult to distinguish without the differentiator neuron features.

	\section{Conclusion and Discussion}
	
	In this work, we developed a comprehensive human-inspired multimodal perception system implemented on a soft robotic hand. The seamless integration between the multimodal data stream and the state-of-the-art performance in object and pose recognition tasks demonstrate the feasibility and advantages of using the human-inspired neuromorphic framework in multimodal robotic perception.
		
	To emulate the multimodal sensory feedback of the human hand, we integrated the soft anthropomorphic robotic hand with sensors corresponding to the human receptors and utilized the receptor models to convert the sensory signals at different rates to sparse spike trains. The neuromorphic signals are processed by a five-layered SNN trained with the SLAYER algorithm. The human receptor-inspired encoding scheme can facilitate the implementation of the method on devices that directly interact with humans, such as the cognitive prosthesis that provides feedback to users. The neuromorphic implementation allows it to be deployed on low-power neuromorphic platforms. 
		
	We evaluated the performance of the proposed methods in object and pose recognition tasks. The results show that the method can accurately recognize objects from the YCB benchmark using grasping interactions. The 99.5\% accuracy in the 14-class object recognition task significantly outperforms the baseline ANN models and the accuracy in similar tasks in previous works. Furthermore, in the more challenging 37-class object and pose recognition task, we also reached 97.14\% accuracy, enabling further exploration or manipulation based on the current pose of the objects without visual information. The result of the ablation tests confirms that integrating multiple sensory modalities can improve recognition performance compared to using individual sensors as in previous works.
		
	In material classification tasks based on thermal property, the proposed differentiator neuron effectively contributes to the performance, enabling the sensorized hand to distinguish all four materials accurately. The proposed pipeline enables the soft anthropomorphic hand to interact more versatilely and efficiently with the environment, utilizing multimodal sensory signals as the human hand.
	
	\section{Acknowledgment}
	This work was supported by the The German Aerospace Center (DLR) under Grant 01GQ2108 and European Research Council (ERC) under Grant 101098308.
		

	\bibliographystyle{IEEEtran}
	\bibliography{ref}

\end{document}